\documentclass[11pt,a4paper]{article}
\usepackage[hyperref]{acl2020}
\pdfoutput=1
\usepackage{times}
\usepackage{latexsym}
\usepackage{amsmath}
\usepackage{tabularx}
\usepackage{graphicx}
\usepackage{url}

\usepackage{microtype}

\aclfinalcopy % Uncomment this line for the final submission
\title{MTLHealth: A Deep Learning System for Detecting Disturbing Content in Student Essays}

\author{Joseph Valencia \\
  ACT, Inc. \\
  Oregon State University\\
  \texttt{valejose@oregonstate.edu} \\\And
  Erin Yao \\
  ACT, Inc. \\
  \texttt{erin.yao@act.org} \\}

\date{07/31/2019}

\begin{document}

\maketitle
\begin{abstract}
    Essay submissions to standardized tests like the ACT occasionally include references to bullying, self-harm, violence, and other forms of disturbing content. Graders must take great care to identify cases like these and decide whether to alert authorities on behalf of students who may be in danger. There is a growing need for robust computer systems to support human decision-makers by automatically flagging potential instances of disturbing content. This paper describes MTLHealth, a disturbing content detection pipeline built around recent advances from computational linguistics, particularly pre-trained language model Transformer networks.
\end{abstract}

\section{Introduction}

In the United States and many other developed countries, mental health crises are becoming more and more visible, including within the youth population. Suicides among 15-19 year-olds are on the rise \citep{cdcmmwr_quickstats:_2017}. Mental health issues place a huge strain on students and their families, as well as on teachers, social workers, and public health officials. 

ACT, Inc. is an educational non-profit that provides a variety of testing and learning products. It is well known for providing assessments for college admissions, k-12 subject areas, workforce skills and social emotional learning skills. Machine learning and natural language processing technology play a vital role in the organization's ability to efficiently process these examinations, including in automated scoring engines to evaluate short and extended constructed response questions in partnership with human graders.

Occasionally, issues related to mental health and trauma surface in essay-based examinations. Graders who are tasked with evaluating essays sometimes notice disturbing content that could necessitate intervention from authorities.

In the same way that machine learning and natural language processing techniques help to grade essays, they can help educational organizations like ACT monitor submissions for disturbing content. Recent advances in deep learning for natural language processing (NLP) have brought great improvements in general language understanding tasks. Such systems can be adapted to recognize language indicative of mental health issues and trauma.

This paper describes a study conducted at ACT with the goal of building a robust deep learning model to automatically flag disturbing content in student responses. Based in part on previous definitions, ACT considers disturbing content to include references to: (1) Attempts at or thoughts of suicide/self-harm, (2) violence or plans to harm others, (3) Physical, sexual, or emotional abuse (4), bullying or social isolation, and (5) substance abuse \citep{burkhardt_morales_lottridge_wood_2017} \citep{Ormerod2018NeuralNA}. 

The resulting system is called MTLHealth. MTLHealth was designed to target student constructed responses, but its predictions could be useful across a wide variety of applications and textual domains, such as for monitoring social media. Note that further evaluation studies should be conducted to understand the performance of the system before it is used operationally.

\section{Related Work}

Prior research provided a strong foundation for the development of MTLHealth. Two areas of the computer science literature proved especially relevant: general methods in natural language understanding, and NLP applications in computational psychology. Additionally, a few papers have explored the specific problem of machine learning approaches to detecting disturbing content.

\subsection{Deep Learning for Natural Language Understanding}

Particularly in the period of 2018-2019, a wave of papers introduced systems that leveraged new approaches to transfer learning, a paradigm in which models trained on one type of data are re-purposed for a new task. These systems used unsupervised pre-training methods to initialize models for a variety of downstream tasks.

The power of transfer learning is apparent in its successful application to a number of challenging benchmark tests. SQuAD is one such challenge, in which the answer to a question must be highlighted within a context paragraph \citep{wang_glue:_2018}. GLUE is another benchmark, consisting of nine diverse language understanding tasks, including natural language inference, sentiment analysis, and paraphrase generation \citep{rajpurkar_know_2018}. On both datasets, deep learning systems have matched human baseline performance.

Two overarching trends in the literature appear to be responsible for much of the progress in language understanding. One is the use of pre-training on massive datasets using objectives similar to language modeling. The other is an increased reliance on Transformer neural network layers.

\subsubsection{Language Model Pre-Training}

Language modeling is the task of inferring a word in a sentence from the previous words in the sentence. Given the text, `` School and work have been pretty stressful lately, I really need a \textit{BLANK} '', a well-trained language model should assign high probability to appropriate words like `vacation', or `nap' and low probability to nonsensical concluding words like `dolphin' or `running'. 

The problem formulation of language modeling is simple, but successful performance requires a complex model of the statistical properties of text. Written language exhibits rich structure -- sentences must adhere to grammatical and other syntactic rules, and the co-occurrence of words captures a good deal of their semantic relationships \citep{linzen_assessing_2016}. Additionally, language modeling is an unsupervised method, meaning it requires no tedious annotation of data by humans. Moreover, the Internet provides easy access to vast amounts of English text to train on. Because of these desireable qualities, language modeling is increasingly recognized as the ideal task for pre-training NLP models \citep{howard_universal_2018}.

Older methods in NLP produced static vector representations of words through training tasks similar to language modeling. For example, Word2Vec calculates a probability distribution along a sliding window of text to build a representation of words as dense vectors of small floating point values \citep{mikolov_distributed_2013}. Similar words have similar representations in vector space and simple arithmetic operators like addition and subtraction can often approximate simple analogies, for example:
$$ \vec{King} - \vec{Man} +\vec{Woman} \approx \vec{Queen} $$

ELMo is one of the pioneering systems that used deep learning language model pre-training to significantly approve performance across a variety of downstream tasks \citep{peters_deep_2018}. ELMo uses a bi-LSTM architecture where one LSTM is trained as an ordinary left-to-right language model and another as a right-to-left language model. Each token in a sentence is represented as a function of all cell states from both LSTMs. This representation provides more context-specific information than does a static word vector.

ULMFit is a language model approach to text classification based on `fine-tuning', which consists of transferring a pre-trained network to a separate task and re-training the entire network on data for this downstream task \citep{howard_universal_2018}. It was the first model to propose a framework for generalized fine-tuning, which was a departure from approaches to transfer learning that exported fixed representations between tasks.  It also introduces new techniques to vary learning rates and to prevent catastrophic forgetting of previously learned information by gradually unfreezing parameters.

\subsubsection{Transformer Networks}

The Transformer is a neural network architecture first proposed by Google Research in the influential paper ``Attention is All You Need'' \citep{vaswani_attention_2017}. The motivation for the design was the prevalence of recurrent architectures like LSTMs and their relative computational inefficiency. Recurrent neural networks require one computational pass for each word in an input sentence. These must proceed in sequence, so they cannot be readily parallelized. This limitation makes recurrent networks expensive to train on large datasets.

The central innovation of Transformers is to replace recurrence with extensive use of attention mechanisms. Attention mechanisms are learned weights that allow the model to emphasize or `attend to' important parts of a sequence for a particular task. They have proven very useful in applications like machine translation, in which they help to model soft alignments between words in the source and target languages. The paper introduces a technique called self-attention, which compares a sentence with itself, calculating an attention distribution relative to each token to capture the relationships between words. The Transformer is built of sub-modules that calculate multiple self-attention distributions and pass them to a feed-forward layer. These sub-modules are stacked into a large encoder-decoder network. 

BERT uses a large Transformer trained on a masked language model objective \citep{devlin_bert:_2018}. Random tokens are replaced with a special masking token, and the system is trained to recover the masked tokens. This allows BERT to incorporate bi-directional context into the representations that it learns. After this pre-training step, BERT is fine-tuned to specific datasets and significantly advanced the state of the art on GLUE and SQuAD.

Other Transformer based models have since exceed BERT's performances, with improvements ranging from very slight to substantial gains on benchmark metrics. MT-DNN is a system that reproduced the BERT architecture and added multi-task learning, i.e. simultaneous training on several downstream tasks\citep{liu_multi-task_2019}. XLNet is another system that uses an autoregressive formulation of language modeling that removes the need for input masking \citep{yang_xlnet:_2019}. RoBERTa is a replication of BERT that better tunes its hyper-parameters \citep{liu_roberta:_2019}.

\subsection{Computational Psychology}

An emerging sub-field of research within computational psychology involves applying general NLP techniques to analyze the connection between language use and mental health. Many studies make use of textual data from web sources, particularly from Twitter and Reddit. 

The Association for Computational Linguistics holds a yearly workshop in Computational Linguistics and Clinical Psychology (CLPsych). CLPysch has sponsored the creation of multiple mental health datasets and held competitions on machine learning for mental health. Past competitions have centered around efforts to triage posts on the suicide support website ReachOut.org and to predict depression and anxiety levels throughout life based on youth essays \citep{milne_clpsych_2016} \citep{lynn_clpsych_2018}. The most recent involved predicting the degree of suicide risk on the r/SuicideWatch subreddit \citep{zirikly_clpsych_2019}.

A paper by Gkotsis demonstrated the ability of simple fully-connected and convolutional neural network architectures to classify Reddit posts as originating from mental health subreddits vs non mental health subreddits \citep{gkotsis_characterisation_2017}. A study by Yates used similar methods on user posts in non mental-health subreddits to predict whether those same users would later announce a depression diagnosis on r/depression \citep{yates_depression_2017}.

\subsection{Disturbing Content Detection}

Some prior work has directly explored the problem of detecting disturbing content in student responses. Earlier research at ACT produced a disturbing content pipeline based on an ensemble of multiple non-neural machine learning methods trained on selected Reddit posts \citep{burkhardt_morales_lottridge_wood_2017}. A study by the American Institutes for Research built a classifier for a large internal dataset of constructed response and compared the performance of several varieties of recurrent neural network architectures. \citep{Ormerod2018NeuralNA}.

\section{Methods}

\begin{figure*}

  \centering \includegraphics[width=\textwidth]{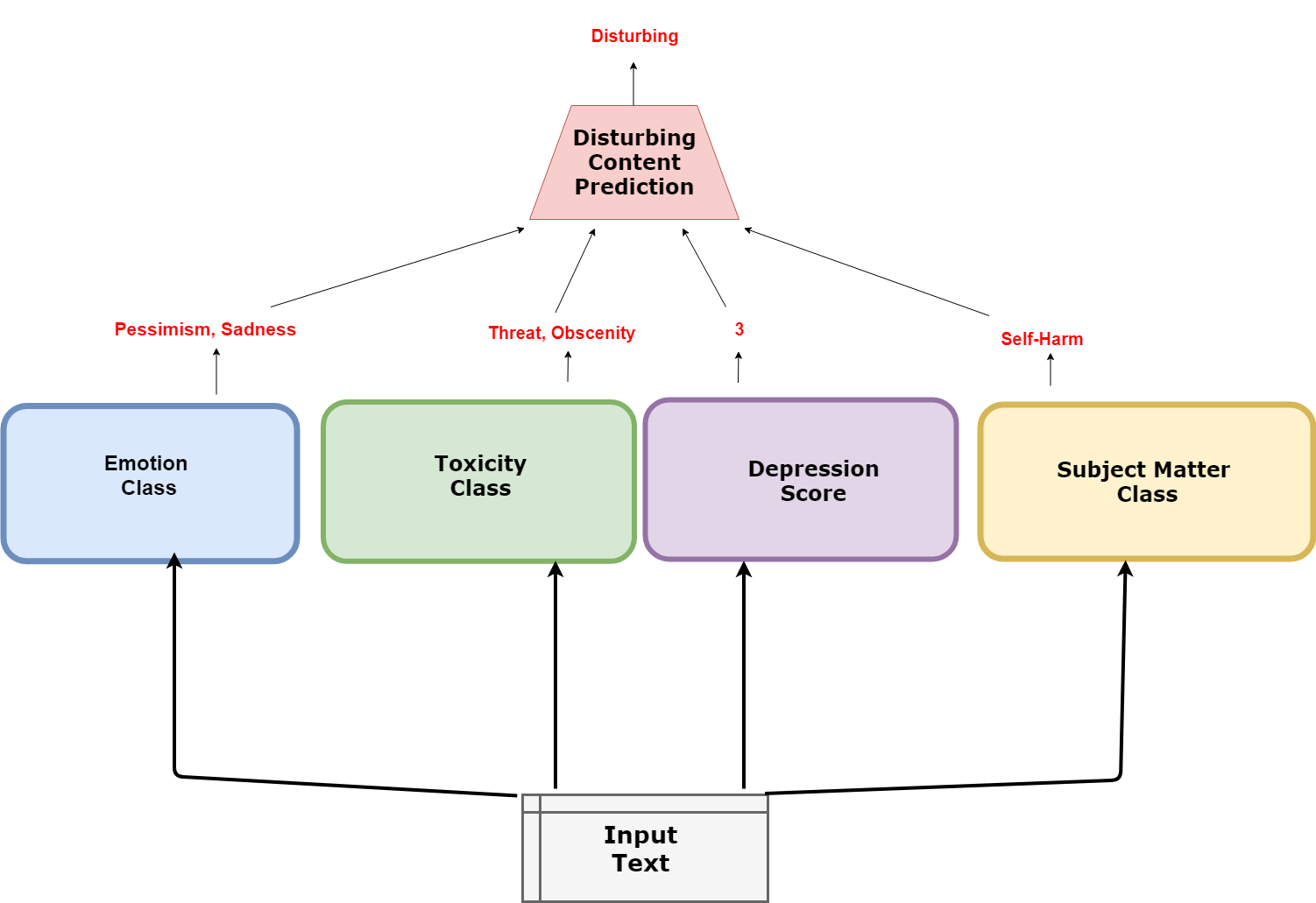}
  \caption{Pipeline Architecture for MTLHealth. }
  \label{fig:architecture}
\end{figure*}

The basic architecture of MTLHealth consists of BERT models trained on various tasks related to mental health and disturbing content. To build a classifier for disturbing content, the prediction scores obtained from evaluating student responses serve as features for a logistic regression classifier. Figure \ref{fig:architecture} is an illustration of this pipeline.

\subsection{Datasets}

MTLHealth is trained on a variety of labeled datasets as described below.

\begin{itemize}
    \item \textit{Emotion Classification}: $\sim$6k Tweets annotated with 11 emotions including anger, disgust, joy, sadness, and trust.
    \citep{mohammad_semeval-2018_2018}
    \item \textit{Toxic Comment Classification}: $\sim$150k Wikipedia comments annotated with labels including toxic, hate speech, obscene, and threat.\footnote{\url{https://www.kaggle.com/c/jigsaw-toxic-comment-classification-\\challenge}}
    \item \textit{UK Birth Cohort Essays}: $\sim$10k essay responses from 11 year olds in 1969 who were asked to imagine their lives at age 25. Each essay is annotated with a numerical score of its author's depression level as measured by a psychological examination following the British Social Adjustment Guide.
    \citep{university_college_london_national_2018} \citep{zirikly_clpsych_2019}
    \item \textit{Subreddit Data}: $\sim$357k user posts in Reddit forums focusing on issues related to mental health along with control posts from subreddits unrelated to mental health. Annotated with subreddit origin information. Some examples are: r/StopSelfHarm, r/SuicideWatch, r/addiction, and control.\footnote{ Mental health data  obtained via personal correspondence with author of \citep{gkotsis_characterisation_2017} and control taken from Reddit dump at \url{https://files.pushshift.io/reddit/comments/}.}
    \item \textit{ACT Constructed Responses}: Consists of 102 constructed responses annotated with disturbing status. 42 flagged as disturbing (label of 1) , 80 flagged as normal (label of 0).
\end{itemize}

The presence of disturbing content in student responses is a relatively rare occurrence. To bolster the MTLHealth system, non ACT datasets were used to encourage the system to flag a wider range of disturbing responses.

The selected datasets were chosen to ensure the system is informed by data that is diverse in terms of domain (short tweets vs long Reddit posts), age of speakers (children vs adults), time period (1960's vs present day), and topic. Emotion and toxicity are somewhat distinct from mental health, but may be strongly correlated with mental health outcomes \citep{hu_relation_2014}. 

\subsection{Training Setup}

The non-profit AI organization HuggingFace maintains pytorch-transformers, a library for the deep learning framework PyTorch that allows users to build models initialized with pre-trained weights from Google \footnote{\url{https://github.com/huggingface/pytorch-transformers}}. MTLHealth uses BERT-Base uncased. BERT-Base uncased is the smaller version of BERT, with 12 Transformer layers, a hidden dimension of 768, and nearly 110M total parameters \citep{devlin_bert:_2018}.

Following the usage guide for BERT, MTLHealth pre-pends all input sentences with the special token `[CLS]'. After fine-tuning, the vector output corresponding to this token serves as the overall representation of the sentence. The pooled output is passed through multiple fully-connected layers with ReLU functions  Eq. \eqref{eq:relu} before being passed to a task-specific output layer.

Training neural networks involves selecting hyper-parameters. 

The `learning rate' is a multiplier for the parameter updates which are calculated during backpropagation. This value affects convergence of the model, if the learning rate is set too low, training could take excessively long. If the learning rate is too high, training may be unstable and the system may never converge. MTLHealth uses $lr = 2e-5$.

The `maximum sequence length' defines a maximum number of tokens beyond which input sentences are truncated. Unlike many training schemes, MTLHealth does not use a fixed batch size. Instead, it accepts a hyper-parameter of `maximum tokens per batch' and attempts to batch sequences of similar sequence length to be close to this value. This scheme follows the implementation found in \citep{rush-2018-annotated}. 

MTLHealth makes use of a technique called dropout, where a percentage of the output is randomly set to zero, which helps to prevent over-fitting by removing noise in the representations \citep{srivastava_dropout:_nodate}. Dropout probability is set at $p = 0.5$.

During fine-tuning for each task, all layers, including those belonging to the pre-trained model, are trained using the `Adam' optimizer, a variation of the standard stochastic gradient descent algorithm that incorporates the concept of `momentum' to automatically scale the learning rate \citep{kingma_adam:_2014}.

Training typically involves multiple passes through the data, which is known as an epoch. Training is allowed to continue for a maximum number of training epochs $max\_epochs=20$. If a chosen task-specific evaluation metric does not improve for several consecutive epochs $patience=5$, training stops automatically.

Hyper-parameter values not found here are listed in the Appendix.

There are three basic problem formulations that MTLHealth is trained on: classification, multi-label classification, and regression. A BERT model was trained for each dataset with an appropriate problem formulation:
classification for the Reddit dataset, multi-label
classification for the Twitter Emotion and
Wikipedia Toxicity datasets, and regression for the
1969 UK Essay dataset.

\begin{itemize}
    \item \textit{Classification} involves identifying a sentence as belonging to one of $C$ classes. Classes are mutually exclusive. The output activation function is a logistic softmax. Eq. \eqref{eq:softmax}. Classification is optimized with the cross-entropy loss function. Eq. \eqref{eq:crossentropy}.
    \item \textit{Multi-label classification} involves identifying a sentence as belonging to 0-C of $C$ classes. Classes are not mutually exclusive, and a sentence can potentially carry some, all or none of the class labels. The output activation function is a sigmoid. Eq. \eqref{eq:sigmoid}. Multi-label classification is optimized with the cross-entropy loss function . Eq. \eqref{eq:crossentropy}.
    \item \textit{Regression} involves predicting a numerical score for a sentence. The output activation function is an identity. Regression is optimized with the mean squared-error loss function Eq. \eqref{eq:mse}. 
\end{itemize}

After training, each BERT model can be applied to any input text. For regression, the output is a predicted score. For classification (including multi-label), the output is the probability of belonging to each class. In turn, these predictions serve as input features for a simple logistic regression classifier. This classifier yields a single yes or no label indicating whether the text includes disturbing content. This system is trained to classify the ACT constructed response data.

All training and experiments were conducted on single p2.xlarge AWS instance with access to half of an NVIDIA Tesla K80 GPU with 12 GB RAM.

\section{Results}

\begin{table}[]

\caption{Toxic Comment Multi-label Classification}
\label{tab:toxic_results}
\begin{tabular}{|l|l|l|l|}
\hline
Model                   & ROC-AUC & mic-F1 & mac-F1 \\
\hline  
Crusader          & 0.98856   &    --      &   --       \\
neongen                  & 0.98822      &   --       &    --      \\
Adversarial  & 0.98805      &  --        &    --      \\
MTLHealth               & 0.9824    & 64.3     & 26.3    \\
\hline
\end{tabular}
\end{table}

\begin{table}[]

\caption{Twitter Emotion Multi-label Classification}
\label{tab:emotion_results} 
\begin{tabular}{|l|l|l|l|}
\hline
Model        & Accuracy & F1-mic & F1-mac \\
\hline
NTUA-SLP     & 58.8     & 70.1     & 52.8     \\
TCS Research & 58.2     & 69.3     & 53       \\
PlusEmo2Vec  & 57.6     & 69.2     & 49.7     \\
MTLHealth    & 56.9     & 69.2     & 51.5    \\
\hline
\end{tabular}
\end{table}

\begin{table}

\caption{UK Essay Regression}
\label{tab:uk_results}
\begin{tabular}{|l|l|l|}
\hline
Model           & Pearson R & MAE \\
\hline  
Coltekin et al. & 0.467     & 0.968 \\
UGent-IDLab 1   & 0.454     & 1.004 \\
UKNLP 1         & 0.433     & 0.951 \\
MTLHealth       & 0.288     & 0.500 \\
\hline
\end{tabular}
\end{table}

\begin{table}[]
\caption{Reddit Post Classification}
\label{tab:reddit_results}
\begin{tabular}{|l|l|l|l|} 
\hline
Acc. & Recall & Precision & F1 \\
\hline 
84.4 & 84.0 & 85.2 & 84.6 \\
\hline
\end{tabular}
\end{table}

\begin{table}[]
\caption{ACT Disturbing Content Classification}
\label{tab:act_results} 
\begin{tabular}{|l|l|l|l|l|l|}
\hline
Features &Acc. & Rec. & Prec. & F1 \\
\hline
UK Essays & --    & -- & -- & -- \\
Toxic  &84.4    & 64.4 & 90.5 & 73.3 \\
Emotion  &89.3    & 81.4 & 89.0 & 83.5 \\
Reddit &90.2    & 90.3 & 85.1 & 86.7 \\
Toxic+Emotion &92.6   & 86.1 & 93.1 & 88.6 \\
Redd.+Toxic+Emo. &95.1    & 88.6 & 97.8 & 92.3\\
\hline
\end{tabular}
\end{table}

 MTLHealth was trained for multi-label classification on the Wikipedia Toxic Comment dataset. The train and test datasets were restricted to a subset of about 6k examples in order to limit training time. Table \ref{tab:toxic_results} compares the performance of MTLHealth with the three top models from the Kaggle Leaderboard. The only metric on the leaderboard is ROC-AUC, in which MTLHealth performed comparably to the previous top submissions. Also included are macro-averaged and micro-averaged F1 for MTLHealth. Eq. \eqref{eq:f1}.

 MTLHealth was trained for multi-label classification on the Twitter Emotion Dataset. Table \ref{tab:emotion_results} compares its performance with the three best performing models on Task A from the ACL SemEval 2018 Workshop. \citep{mohammad_semeval-2018_2018}. MTLHealth achieved competitive results in all three metrics, including the main contest metric of  multi-label accuracy (also known as Jaccard Index). Eq. \eqref{eq:jaccard}.  

 MTLHealth was trained for regression on the UK Essay Depression Score dataset. This problem first appeared as Task A of the CLPsych workshop at ACL 2018.  The training and testing datasets were constructed using the original data provided by the UK Data Service. Therefore, the breakdown into test and train sets will be different from that used in the contest. Table \ref{tab:uk_results} compares MTLHealth with the top three models from this contest. MTLHealth performs considerably worse than the others on Disattenuated Pearson R , but much better on the Mean Absolute Error metric. Eq. \eqref{eq:mae} Eq. \eqref{eq:pearson}

\begin{figure*}

  \centering \includegraphics[width=\textwidth]{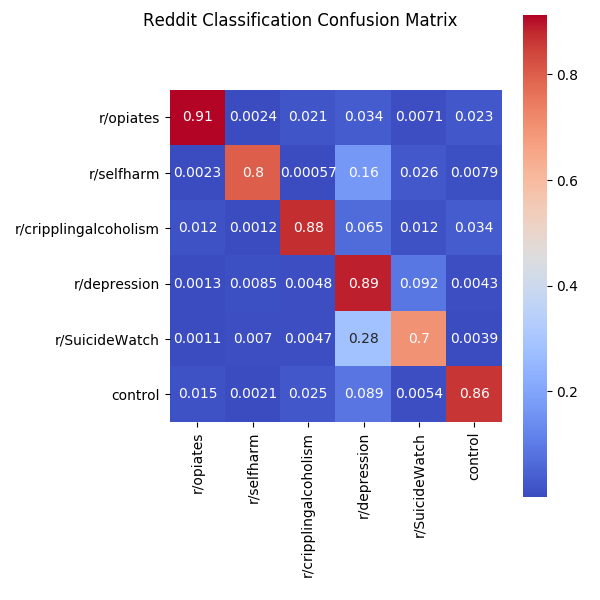}
  \caption{Confusion Matrix for Reddit Classification}
  \label{fig:Reddit}
\end{figure*}

Table \ref{tab:reddit_results} lists the classification metrics for the Reddit task. Because the training dataset adds manually gathered control data to a subset of the mental health data from \citep{gkotsis_characterisation_2017}, MTLHealth is not directly comparable with any prior system. As shown in the confusion matrix in Figure \ref{fig:Reddit}, the system achieves a high rate of correct classifications. By far the largest sources of confusion are misclassification of posts from r/SuicideWatch and r/selfharm as coming from r/depression, which can be explained by the close semantic relationship of these three subreddits.

The final disturbing content predictor for MTLHealth is a basic logistic regression model trained on the ACT constructed response data. Every entry in the constructed response set was split into chunks of 50 tokens and passed chunk by chunk into the BERT models for the Wikipedia Toxic, Twitter Emotion, and Reddit tasks. The feature representations for each constructed response were the output scores averaged over all chunks. The UK Essay model was excluded from the analysis due to its poor performance on testing data. Table \ref{tab:act_results} shows results from a 5-fold cross-validation evaluation using different combinations of the feature sets.

 \begin{figure*}
  \centering \includegraphics[width=\textwidth]{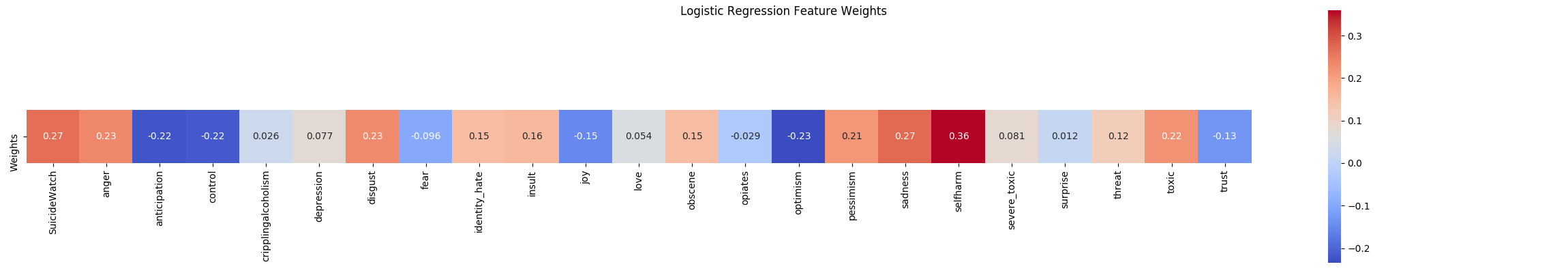}
  \caption{Disturbing Content Classifier Weights}
  \label{fig:feature_weights}
\end{figure*}

\section{Discussion}

Overall, the results of the study indicated that the BERT architecture adapts well to many applications related to disturbing content. On the Reddit, Wiki Toxic, and Twitter Emotion datasets, MTLHealth achieved high performance on all metrics. These features were highly useful in predicting the disturbing content status of ACT constructed responses.

\subsection{Disturbing Content Feature Importance}

Table \ref{tab:act_results} reports the results of an ablation study, which demonstrated the usefulness of the features extracted by the various BERT pipelines. The Reddit model produced by far the most predictive features of any single model, and the toxicity pipeline produced the least useful features. The best results were obtained using all feature sets besides the depression scores from the UK Data Essays. Fig \eqref{fig:feature_weights} represents the magnitude of the feature weights learned by the final logistic regression in MTLHealth. Features like `suicideWatch' ,`severe\_toxic', and `sadness' received high positive weights, meaning they are positive contributions to classifying a response as disturbing. Similarly, features like `optimism','trust', and `control' were highly negative indicators of disturbing content.

\subsection{Comparison to Prior Work}

The use of transfer learning and Transformer architectures allows MTLHealth to benefit from the state-of-the art methods in 2019. The incorporation of diverse training data also makes MTLHealth unique among disturbing content detection systems in its capabilities across a variety of tasks.

MTLHealth was trained on over 200 times more text examples than the prior work at ACT. The older version only uses Reddit sources, while MTLHealth is able to make predictions on varying input domains and problem formations. \citep{burkhardt_morales_lottridge_wood_2017}. Whereas the old system used an ensemble of non-neural machine learning models all trained on one dataset, MTLHealth builds an ensemble that incorporates information from several different datasets into its predictions. This should predispose MTLHealth towards more generally applicable predictions, although this must be confirmed with validation on more data. 

The paper from the American Institutes for Research (AIR) does not report conventional classification metrics, instead focusing on the total number of disturbing essays that are caught by choosing a fixed percentage of the essays with the highest probability of being disturbing \citep{Ormerod2018NeuralNA}. Crucially, unlike the AIR system, most of the training data for MTLHealth comes from publicly available datasets. 

\subsection{Future Work}

The name MTLHealth refers both to its focus on mental health, and to the eventual goal of incorporating a training paradigm known as multi-task learning into the package. A good deal of literature indicates that simultaneously training a system on related tasks using shared parameters can yield better results than training separate models for each individual dataset  \citep{hashimoto_joint_2016} \citep{McCann2018decaNLP}. Future work will extend MTLHealth to an architecture based on MT-DNN \citep{liu_multi-task_2019}. In essence, each task will have a small number of output layers specific to the task, while the lower level representations will be in common for all tasks. This scheme will require the model to learn representations that are widely applicable across tasks.

Additionally, the MTLHealth software must undergo more rigorous review and testing before it is put into operation as part of the disturbing content detection protocol for ACT. Specifically, it should conduct more extensive empirical validation of MTLHealth predictions as additional constructed responses are annotated and transcribed. Given enough of this internal data, MTLHealth could potentially be trained to classify constructed responses in an end-to-end manner, i.e. without the intermediate feature extractors, or with the student data as part of an ensemble with the existing trained models. It must also consider the availability of manual graders in order to decide a threshold above which a probability estimate from MTLHealth should trigger an alert.

\section*{Acknowledgments}

The authors would like to acknowledge John Whitmer and Brian LaMure of ACT for providing access to essential computing resources. \\

\bibliographystyle{acl_natbib}
\bibliography{acl2020}

\begin{thebibliography}{27}
\expandafter\ifx\csname natexlab\endcsname\relax\def\natexlab#1{#1}\fi

\bibitem[{Burkhardt et~al.(2017)Burkhardt, Morales, Lottridge, and
  Wood}]{burkhardt_morales_lottridge_wood_2017}
Amy Burkhardt, Carlo Morales, Susan Lottridge, and Scott Wood. 2017.
\newblock \href
  {https://www.slideshare.net/AmyBurkhardt1/the-automatic-detection-of-disturbing-content-within-online-tests-ncme-2017final}
  {[link]}.

\bibitem[{CDCMMWR(2017)}]{cdcmmwr_quickstats:_2017}
CDCMMWR. 2017.
\newblock \href {https://doi.org/10.15585/mmwr.mm6630a6} {{QuickStats}:
  {Suicide} {Rates} for {Teens} {Aged} 15–19 {Years}, by {Sex} — {United}
  {States}, 1975–2015}.
\newblock \emph{MMWR. Morbidity and Mortality Weekly Report}, 66.

\bibitem[{Devlin et~al.(2018)Devlin, Chang, Lee, and
  Toutanova}]{devlin_bert:_2018}
Jacob Devlin, Ming-Wei Chang, Kenton Lee, and Kristina Toutanova. 2018.
\newblock \href {http://arxiv.org/abs/1810.04805} {{BERT}: {Pre}-training of
  {Deep} {Bidirectional} {Transformers} for {Language} {Understanding}}.
\newblock \emph{arXiv:1810.04805 [cs]}.
\newblock ArXiv: 1810.04805.

\bibitem[{Gkotsis et~al.(2017)Gkotsis, Oellrich, Velupillai, Liakata, Hubbard,
  Dobson, and Dutta}]{gkotsis_characterisation_2017}
George Gkotsis, Anika Oellrich, Sumithra Velupillai, Maria Liakata, Tim J.~P.
  Hubbard, Richard J.~B. Dobson, and Rina Dutta. 2017.
\newblock \href {https://doi.org/10.1038/srep45141} {Characterisation of mental
  health conditions in social media using {Informed} {Deep} {Learning}}.
\newblock \emph{Scientific Reports}, 7:45141.

\bibitem[{Hashimoto et~al.(2016)Hashimoto, Xiong, Tsuruoka, and
  Socher}]{hashimoto_joint_2016}
Kazuma Hashimoto, Caiming Xiong, Yoshimasa Tsuruoka, and Richard Socher. 2016.
\newblock \href {http://arxiv.org/abs/1611.01587} {A {Joint} {Many}-{Task}
  {Model}: {Growing} a {Neural} {Network} for {Multiple} {NLP} {Tasks}}.
\newblock \emph{arXiv:1611.01587 [cs]}.
\newblock ArXiv: 1611.01587.

\bibitem[{Howard and Ruder(2018)}]{howard_universal_2018}
Jeremy Howard and Sebastian Ruder. 2018.
\newblock \href {http://arxiv.org/abs/1801.06146} {Universal {Language} {Model}
  {Fine}-tuning for {Text} {Classification}}.
\newblock \emph{arXiv:1801.06146 [cs, stat]}.
\newblock ArXiv: 1801.06146.

\bibitem[{Hu et~al.(2014)Hu, Zhang, Wang, Mistry, Ran, and
  Wang}]{hu_relation_2014}
Tianqiang Hu, Dajun Zhang, Jinliang Wang, Ritesh Mistry, Guangming Ran, and
  Xinqiang Wang. 2014.
\newblock \href {https://doi.org/10.2466/03.20.PR0.114k22w4} {Relation between
  {Emotion} {Regulation} and {Mental} {Health}: {A} {Meta}-{Analysis}
  {Review}}.
\newblock \emph{Psychological Reports}, 114(2):341--362.

\bibitem[{Kingma and Ba(2014)}]{kingma_adam:_2014}
Diederik~P. Kingma and Jimmy Ba. 2014.
\newblock \href {http://arxiv.org/abs/1412.6980} {Adam: {A} {Method} for
  {Stochastic} {Optimization}}.
\newblock \emph{arXiv:1412.6980 [cs]}.
\newblock ArXiv: 1412.6980.

\bibitem[{Linzen et~al.(2016)Linzen, Dupoux, and
  Goldberg}]{linzen_assessing_2016}
Tal Linzen, Emmanuel Dupoux, and Yoav Goldberg. 2016.
\newblock \href {http://arxiv.org/abs/1611.01368} {Assessing the {Ability} of
  {LSTMs} to {Learn} {Syntax}-{Sensitive} {Dependencies}}.
\newblock \emph{arXiv:1611.01368 [cs]}.
\newblock ArXiv: 1611.01368.

\bibitem[{Liu et~al.(2019{\natexlab{a}})Liu, He, Chen, and
  Gao}]{liu_multi-task_2019}
Xiaodong Liu, Pengcheng He, Weizhu Chen, and Jianfeng Gao. 2019{\natexlab{a}}.
\newblock \href {http://arxiv.org/abs/1901.11504} {Multi-{Task} {Deep} {Neural}
  {Networks} for {Natural} {Language} {Understanding}}.
\newblock \emph{arXiv:1901.11504 [cs]}.
\newblock ArXiv: 1901.11504.

\bibitem[{Liu et~al.(2019{\natexlab{b}})Liu, Ott, Goyal, Du, Joshi, Chen, Levy,
  Lewis, Zettlemoyer, and Stoyanov}]{liu_roberta:_2019}
Yinhan Liu, Myle Ott, Naman Goyal, Jingfei Du, Mandar Joshi, Danqi Chen, Omer
  Levy, Mike Lewis, Luke Zettlemoyer, and Veselin Stoyanov. 2019{\natexlab{b}}.
\newblock \href {http://arxiv.org/abs/1907.11692} {{RoBERTa}: {A} {Robustly}
  {Optimized} {BERT} {Pretraining} {Approach}}.
\newblock \emph{arXiv:1907.11692 [cs]}.
\newblock ArXiv: 1907.11692.

\bibitem[{Lynn et~al.(2018)Lynn, Goodman, Niederhoffer, Loveys, Resnik, and
  Schwartz}]{lynn_clpsych_2018}
Veronica Lynn, Alissa Goodman, Kate Niederhoffer, Kate Loveys, Philip Resnik,
  and H.~Andrew Schwartz. 2018.
\newblock \href {https://doi.org/10.18653/v1/W18-0604} {{CLPsych} 2018 {Shared}
  {Task}: {Predicting} {Current} and {Future} {Psychological} {Health} from
  {Childhood} {Essays}}.
\newblock In \emph{Proceedings of the {Fifth} {Workshop} on {Computational}
  {Linguistics} and {Clinical} {Psychology}: {From} {Keyboard} to {Clinic}},
  pages 37--46, New Orleans, LA. Association for Computational Linguistics.

\bibitem[{McCann et~al.(2018)McCann, Keskar, Xiong, and
  Socher}]{McCann2018decaNLP}
Bryan McCann, Nitish~Shirish Keskar, Caiming Xiong, and Richard Socher. 2018.
\newblock The natural language decathlon: Multitask learning as question
  answering.
\newblock \emph{arXiv preprint arXiv:1806.08730}.

\bibitem[{Mikolov et~al.(2013)Mikolov, Sutskever, Chen, Corrado, and
  Dean}]{mikolov_distributed_2013}
Tomas Mikolov, Ilya Sutskever, Kai Chen, Greg~S Corrado, and Jeff Dean. 2013.
\newblock \href
  {http://papers.nips.cc/paper/5021-distributed-representations-of-words-and-phrases-and-their-compositionality.pdf}
  {Distributed {Representations} of {Words} and {Phrases} and their
  {Compositionality}}.
\newblock In C.~J.~C. Burges, L.~Bottou, M.~Welling, Z.~Ghahramani, and K.~Q.
  Weinberger, editors, \emph{Advances in {Neural} {Information} {Processing}
  {Systems} 26}, pages 3111--3119. Curran Associates, Inc.

\bibitem[{Milne et~al.(2016)Milne, Pink, Hachey, and
  Calvo}]{milne_clpsych_2016}
David~N. Milne, Glen Pink, Ben Hachey, and Rafael~A. Calvo. 2016.
\newblock \href {https://doi.org/10.18653/v1/W16-0312} {{CLPsych} 2016 {Shared}
  {Task}: {Triaging} content in online peer-support forums}.
\newblock In \emph{Proceedings of the {Third} {Workshop} on {Computational}
  {Linguistics} and {Clinical} {Psychology}}, pages 118--127, San Diego, CA,
  USA. Association for Computational Linguistics.

\bibitem[{Mohammad et~al.(2018)Mohammad, Bravo-Marquez, Salameh, and
  Kiritchenko}]{mohammad_semeval-2018_2018}
Saif Mohammad, Felipe Bravo-Marquez, Mohammad Salameh, and Svetlana
  Kiritchenko. 2018.
\newblock \href {https://doi.org/10.18653/v1/S18-1001} {{SemEval}-2018 {Task}
  1: {Affect} in {Tweets}}.
\newblock In \emph{Proceedings of {The} 12th {International} {Workshop} on
  {Semantic} {Evaluation}}, pages 1--17, New Orleans, Louisiana. Association
  for Computational Linguistics.

\bibitem[{Ormerod and Harris(2018)}]{Ormerod2018NeuralNA}
Christopher~M. Ormerod and Amy~E. Harris. 2018.
\newblock Neural network approach to classifying alarming student responses to
  online assessment.
\newblock \emph{ArXiv}, abs/1809.08899.

\bibitem[{Peters et~al.(2018)Peters, Neumann, Iyyer, Gardner, Clark, Lee, and
  Zettlemoyer}]{peters_deep_2018}
Matthew~E. Peters, Mark Neumann, Mohit Iyyer, Matt Gardner, Christopher Clark,
  Kenton Lee, and Luke Zettlemoyer. 2018.
\newblock \href {http://arxiv.org/abs/1802.05365} {Deep contextualized word
  representations}.
\newblock \emph{arXiv:1802.05365 [cs]}.
\newblock ArXiv: 1802.05365.

\bibitem[{Rajpurkar et~al.(2018)Rajpurkar, Jia, and
  Liang}]{rajpurkar_know_2018}
Pranav Rajpurkar, Robin Jia, and Percy Liang. 2018.
\newblock \href {http://arxiv.org/abs/1806.03822} {Know {What} {You} {Don}'t
  {Know}: {Unanswerable} {Questions} for {SQuAD}}.
\newblock \emph{arXiv:1806.03822 [cs]}.
\newblock ArXiv: 1806.03822.

\bibitem[{Rush(2018)}]{rush-2018-annotated}
Alexander Rush. 2018.
\newblock \href {https://doi.org/10.18653/v1/W18-2509} {The annotated
  transformer}.
\newblock In \emph{Proceedings of Workshop for {NLP} Open Source Software
  ({NLP}-{OSS})}, pages 52--60, Melbourne, Australia. Association for
  Computational Linguistics.

\bibitem[{Srivastava et~al.()Srivastava, Hinton, Krizhevsky, Sutskever, and
  Salakhutdinov}]{srivastava_dropout:_nodate}
Nitish Srivastava, Geoﬀrey Hinton, Alex Krizhevsky, Ilya Sutskever, and
  Ruslan Salakhutdinov.
\newblock Dropout: {A} {Simple} {Way} to {Prevent} {Neural} {Networks} from
  {Overﬁtting}.
\newblock page~30.

\bibitem[{University
  College~London(2018)}]{university_college_london_national_2018}
UCL Institute of~Education University College~London. 2018.
\newblock National {Child} {Development} {Study}: "{Imagine} you are 25"
  {Essays} ({Sweep} 2, {Age} 11), 1969.
\newblock Technical report, UK Data Service.

\bibitem[{Vaswani et~al.(2017)Vaswani, Shazeer, Parmar, Uszkoreit, Jones,
  Gomez, Kaiser, and Polosukhin}]{vaswani_attention_2017}
Ashish Vaswani, Noam Shazeer, Niki Parmar, Jakob Uszkoreit, Llion Jones,
  Aidan~N Gomez, Łukasz Kaiser, and Illia Polosukhin. 2017.
\newblock \href
  {http://papers.nips.cc/paper/7181-attention-is-all-you-need.pdf} {Attention
  is {All} you {Need}}.
\newblock In I.~Guyon, U.~V. Luxburg, S.~Bengio, H.~Wallach, R.~Fergus,
  S.~Vishwanathan, and R.~Garnett, editors, \emph{Advances in {Neural}
  {Information} {Processing} {Systems} 30}, pages 5998--6008. Curran
  Associates, Inc.

\bibitem[{Wang et~al.(2018)Wang, Singh, Michael, Hill, Levy, and
  Bowman}]{wang_glue:_2018}
Alex Wang, Amanpreet Singh, Julian Michael, Felix Hill, Omer Levy, and
  Samuel~R. Bowman. 2018.
\newblock \href {http://arxiv.org/abs/1804.07461} {{GLUE}: {A} {Multi}-{Task}
  {Benchmark} and {Analysis} {Platform} for {Natural} {Language}
  {Understanding}}.
\newblock \emph{arXiv:1804.07461 [cs]}.
\newblock ArXiv: 1804.07461.

\bibitem[{Yang et~al.(2019)Yang, Dai, Yang, Carbonell, Salakhutdinov, and
  Le}]{yang_xlnet:_2019}
Zhilin Yang, Zihang Dai, Yiming Yang, Jaime Carbonell, Ruslan Salakhutdinov,
  and Quoc~V. Le. 2019.
\newblock \href {http://arxiv.org/abs/1906.08237} {{XLNet}: {Generalized}
  {Autoregressive} {Pretraining} for {Language} {Understanding}}.
\newblock \emph{arXiv:1906.08237 [cs]}.
\newblock ArXiv: 1906.08237.

\bibitem[{Yates et~al.(2017)Yates, Cohan, and Goharian}]{yates_depression_2017}
Andrew Yates, Arman Cohan, and Nazli Goharian. 2017.
\newblock \href {https://doi.org/10.18653/v1/D17-1322} {Depression and
  {Self}-{Harm} {Risk} {Assessment} in {Online} {Forums}}.
\newblock In \emph{Proceedings of the 2017 {Conference} on {Empirical}
  {Methods} in {Natural} {Language} {Processing}}, pages 2968--2978,
  Copenhagen, Denmark. Association for Computational Linguistics.

\bibitem[{Zirikly et~al.(2019)Zirikly, Resnik, Uzuner, and
  Hollingshead}]{zirikly_clpsych_2019}
Ayah Zirikly, Philip Resnik, Özlem Uzuner, and Kristy Hollingshead. 2019.
\newblock \href {https://www.aclweb.org/anthology/W19-3003} {{CLPsych} 2019
  {Shared} {Task}: {Predicting} the {Degree} of {Suicide} {Risk} in {Reddit}
  {Posts}}.
\newblock In \emph{Proceedings of the {Sixth} {Workshop} on {Computational}
  {Linguistics} and {Clinical} {Psychology}}, pages 24--33, Minneapolis,
  Minnesota. Association for Computational Linguistics.

\end{thebibliography}

\appendix

\section{Appendices}
\label{sec:appendix}

\subsection{Additional Hyperparameters}
\begin{tabular}{|l|l|l|}
\hline
Model           & MaxTokens/Batch & MaxLen \\
\hline  
UK Essays & 5000     & 224 \\
Wiki Toxic  & 5400     &324 \\
Twitter Emotion        & 5400     & 324 \\
Reddit      & 5000     & 224 \\
\hline
\end{tabular}
\\
\\
MaxTokens/Batch is maximum number of tokens to fit in a batch.\\
MaxLen is the maximum length of text input in tokens.

\subsection{Classification Metrics}
For all equations, N refers to the number of samples, $e$ refers to the base of the natural logarithm, and $\log$ is base 2.

\begin{equation}
\label{eq:recall}
    Recall = \frac{TP}{TP+FN}
\end{equation}

\begin{equation}
\label{eq:precision}
    Precision = \frac{TP}{TP+FP}
\end{equation}

\begin{equation}
\label{eq:f1}
    F1 = 2 \cdot \frac{Precision\cdot Recall}{Precision+Recall}
\end{equation}

\begin{equation}
\label{eq:jaccard}
    JaccardIndex = \frac{1}{N}\sum_{i=1}^{N}\frac{TP_{i}}{TP_{i}+FP_{i}+FN_{i}}
\end{equation}
TP = True Positives \\
FP = False Positives\\
TN = True Negatives\\
FN = False Negatives\\
Subscripts $i$ for \eqref{eq:jaccard} indicate that metric is calculated on each example and averaged.\\

\subsection{Regression Metrics}

\begin{equation}
\label{eq:pearson}
   Pearson_{D}(X,Y) = {\frac {\operatorname {cov} (X,Y)}{\sigma _{X}\sigma _{Y}}} \cdot \frac{1}{\sqrt{0.77\cdot 0.70}}
\end{equation}

\begin{equation}
\label{eq:mae}
   MAE = \frac{1}{N}\sum_{i=1}^{N} (Y_{i} -\hat{Y_{i}})
\end{equation} \\
$\sigma$ is the standard deviation. \\
$cov$ is the covariance.\\
$Y_{i}$ is the predicted score for sample $i$\\
$\hat{Y_{i}}$ is the true score for sample $i$\\
\eqref{eq:pearson} is a `disattenuated' version of Pearson correlation R that accounts for measurement error and re-scales the metric to [-1.362,1.362] , as proposed in \citep{lynn_clpsych_2018}.

\subsection{Activation Functions}

$\mathcal{C}$ refers to the set of classes for a classification problem.

\begin{equation}
\label{eq:sigmoid}
Sigmoid(x) = \frac{1}{1+e^{-x}}
\end{equation}

\begin{equation}
\label{eq:softmax}
LogSoftmax(x) = \log \frac{e^{x_{i}}}{{\sum_{j \in \mathcal{C}} e^{x_{j}}}}
\end{equation}

\begin{equation}
\label{eq:relu}
ReLU(x) = max(0,x)
\end{equation}

\subsection{Loss Functions}
$\mathcal{C}$ is the set of classes.\\
$p_{i}$ is the true distribution for class $i$.\\
$q_{i}$ is the predicted distribution for class $i$.\\

\begin{equation}
\label{eq:mse}
    MSE = \frac{1}{N}\sum_{i=1}^{N} (Y_{i} - \hat{Y_{i}})^{2}
\end{equation}

\begin{equation}
\label{eq:crossentropy}
CrossEntropy =-\sum _{i\in {\mathcal {C}}}{p_{i}\,\log q_{i}}
\end{equation}

\end{document}